\documentclass{article}

     \PassOptionsToPackage{numbers, compress}{natbib}

\usepackage[preprint]{neurips_2021}
\usepackage{multicol}

\usepackage[utf8]{inputenc} 
\usepackage[T1]{fontenc}    
\usepackage{hyperref}       
\usepackage{url}            
\usepackage{booktabs}       
\usepackage{amsfonts}       
\usepackage{nicefrac}       
\usepackage{microtype}      
\usepackage{xcolor}         
\usepackage{mathtools, amsmath, amsthm}
\usepackage{algorithm}
\usepackage{algorithmic}

\newcommand{\name}{ApproxIFER} 
\newcommand{\namespace}{\name{ }}
\newcommand{\deff}{\mbox{$\stackrel{\rm def}{=}$}}
\newcommand{\Aadv}{A_{\text{adv}}} 
\newcommand{\Aavail}{A_{\text{avl}}}

\newtheorem{theorem}{{Theorem}}

\title{ApproxIFER: A Model-Agnostic Approach to Resilient and Robust  Prediction Serving Systems}

 \author{%
   Mahdi Soleymani \\
   EECS Department \\
   University of Michigan-Ann Arbor\\
   \texttt{mahdy@umich.edu} \\
   \And 
   	Ramy E. Ali\\
   ECE Department\\
   University of Southern California (USC)\\
   \texttt{reali@usc.edu} \\
      \And 
  Hessam Mahdavifar\\
EECS Department \\
University of Michigan-Ann Arbor\\
\texttt{hessam@umich.edu} \\
      \And 
   A. Salman Avestimehr \\
   	ECE Department\\
   University of Southern California (USC)\\
   \texttt{avestime@usc.edu} \\
 }

\begin{document}

\maketitle

\begin{abstract}
Due to the surge of cloud-assisted AI services, the problem of designing resilient prediction serving systems that can effectively cope with stragglers/failures and minimize response delays has attracted much interest. 
The common approach for tackling this problem is replication which assigns the same prediction task to multiple workers. This approach, however, is very inefficient and incurs significant resource overheads.
Hence, a learning-based approach known as parity model (ParM) has been recently proposed which learns models that can generate ``parities’’ for a group of predictions in order to reconstruct the predictions of the slow/failed workers.
While this learning-based approach is more resource-efficient than replication, it is tailored to the specific model hosted by the cloud and is particularly suitable for a small number of queries (typically less than four) and tolerating very few (mostly one) number of stragglers. Moreover, ParM does not handle Byzantine adversarial workers. 
We propose a different approach, named Approximate Coded Inference (\name), that does not require training of any parity models, hence it is agnostic to the model hosted by the cloud and can be readily applied to different data domains and model architectures. 
%
Compared with earlier works, \namespace can handle a general number of stragglers and scales significantly better with the  number of queries. 
Furthermore, \namespace is robust against Byzantine workers. 
Our extensive experiments on a large number of datasets and model architectures also show significant accuracy improvement by up to $58\%$ over the parity model approaches.

\end{abstract}

\section{Introduction}
Machine learning as a service (MLaS) paradigms allow incapable clients to outsource their  computationally-demanding tasks such as neural network inference tasks to powerful clouds \cite{Amazon-AWS-AI,Azure-Studio,Google-Cloud-AI,olston2017tensorflow}. More specifically, prediction serving systems host complex machine learning models and respond to the inference queries of the clients by the corresponding predictions with low latency. To ensure a fast response to the different queries in the presence of stragglers, prediction serving systems distribute these queries on multiple worker nodes in the system each having an instance of the deployed model \cite{crankshaw2017clipper}. Such systems often mitigate stragglers through replication which assigns the same task to multiple workers, either proactively or reactively, in order to reduce the tail latency of the computations \cite{suresh2015c3,Apache-Hadoop,ananthanarayanan2013effective,dean2013tail,shah2015redundant,gardner2015reducing,chaubey2015replicated}. Replication-based systems, however, entail significant overhead as a result of assigning the same task to multiple workers. 

Erasure coding is known to be more resource-efficient compared to replication and has been recently leveraged to speed up distributed computing and learning systems \cite{lee2017speeding,dutta2016short,yu2019lagrange,yu2017polynomial,narra2019slack,muralee2020coded,soto2019dual}. The traditional coding-theoretic approaches, known as the coded computing approaches, are usually limited to polynomial computations and require a large number of workers that depends on the desired computation. Hence, such techniques cannot be directly applied in prediction serving systems. 

\begin{figure}[htb!]
    \centering
    \includegraphics[scale=0.5]{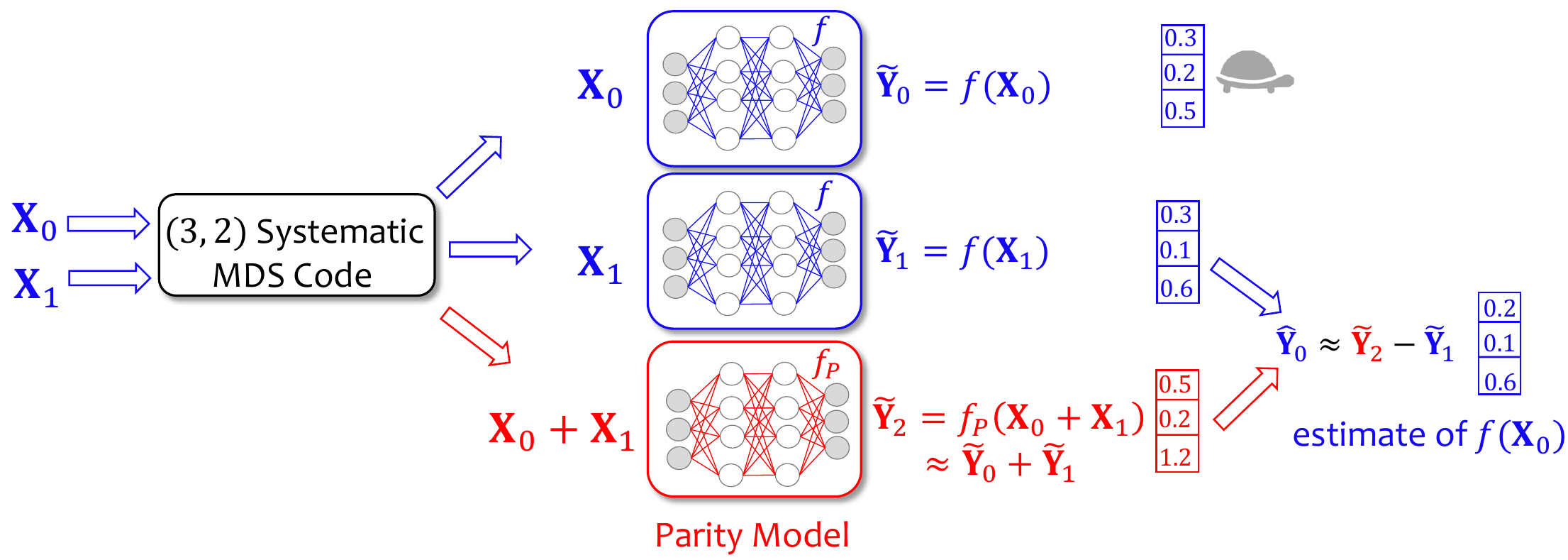}
    \caption{  \footnotesize
An example of ParM is illustrated with $K=2$ queries denoted by $\mathbf X_0$ and $\mathbf X_1$. The goal is to compute the predictions $\mathbf Y_0=f(\mathbf X_0)$ and $\mathbf Y_1=f(\mathbf X_1)$. In this example, the system is designed to tolerate one straggler. Worker $1$ and worker $2$ have the model deployed by the prediction serving system denoted by $f$. Worker $3$, has the parity model $f_P$ which is trained with the ideal goal that  $f_P(\mathbf X_0+\mathbf X_1) = f(\mathbf X_0)+f(\mathbf X_1)$. In this scenario, the first worker is slow and $f(\mathbf X_0)$ is estimated from $f(\mathbf X_1)$ and $f_P(\mathbf X_0+\mathbf X_1)$.}
    \label{fig:ParMExample}
\end{figure}

To overcome the limitations of the traditional coding-theoretic approaches, a learning-based approach known as ParM has been proposed in \cite{kosaian2019parity}. In this approach, the prediction queries are first encoded using an erasure code. These coded queries are then transformed into coded predictions by learning \emph{parity models} to provide straggler resiliency. The desired predictions can be then reconstructed from the fastest workers as shown in Fig. \ref{fig:ParMExample}. By doing this, ParM can be applied to non-polynomial computations with a number of workers that is independent of the computations.

\begin{figure}[htb!]
    \centering
    \includegraphics[scale=0.45]{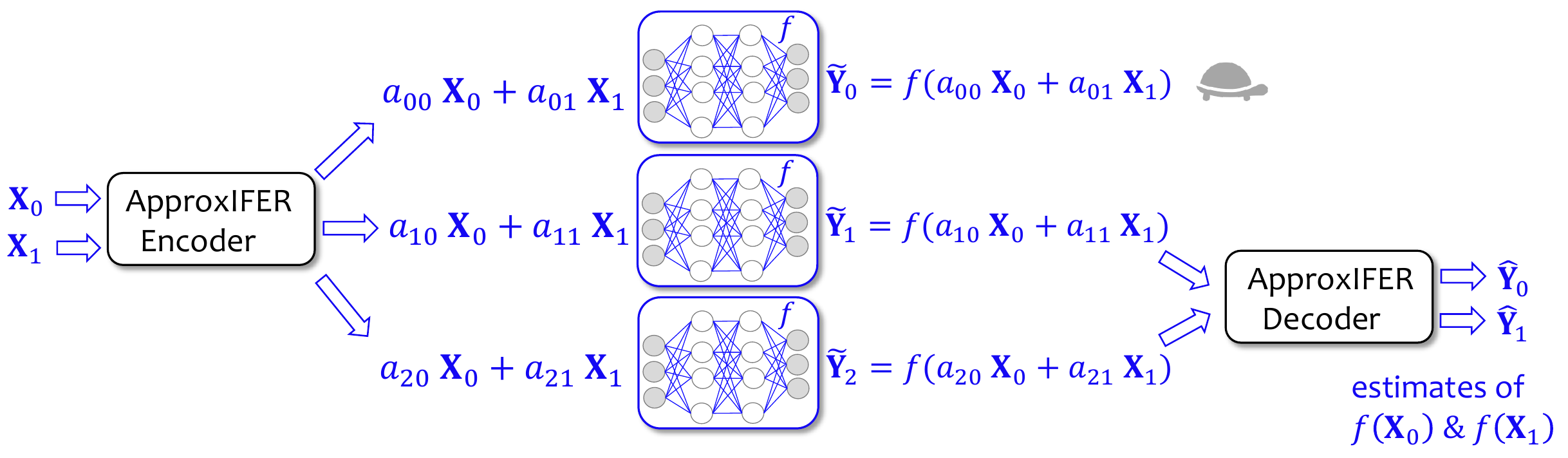}
    \caption{\footnotesize An example of \namespace is illustrated with $K=2$ queries and $S=1$ straggler. Unlike ParM, all workers in \namespace have the same model $f$ which is the model hosted by the cloud. In this scenario, the first worker is slow and $f(\mathbf X_0)$ and $f(\mathbf X_1)$ are estimated from $\tilde{\mathbf Y}_1$ and $\tilde{\mathbf Y}_2$. The key idea of \namespace is that it carefully chooses the coefficients while encoding the queries such that it can estimate the desired predictions from the coded predictions of the fast workers through interpolation.}
    \label{fig:ApproxIferExample}
\end{figure}

\begin{figure}
\centering
\includegraphics[height=0.145\paperheight, width=0.45\linewidth]{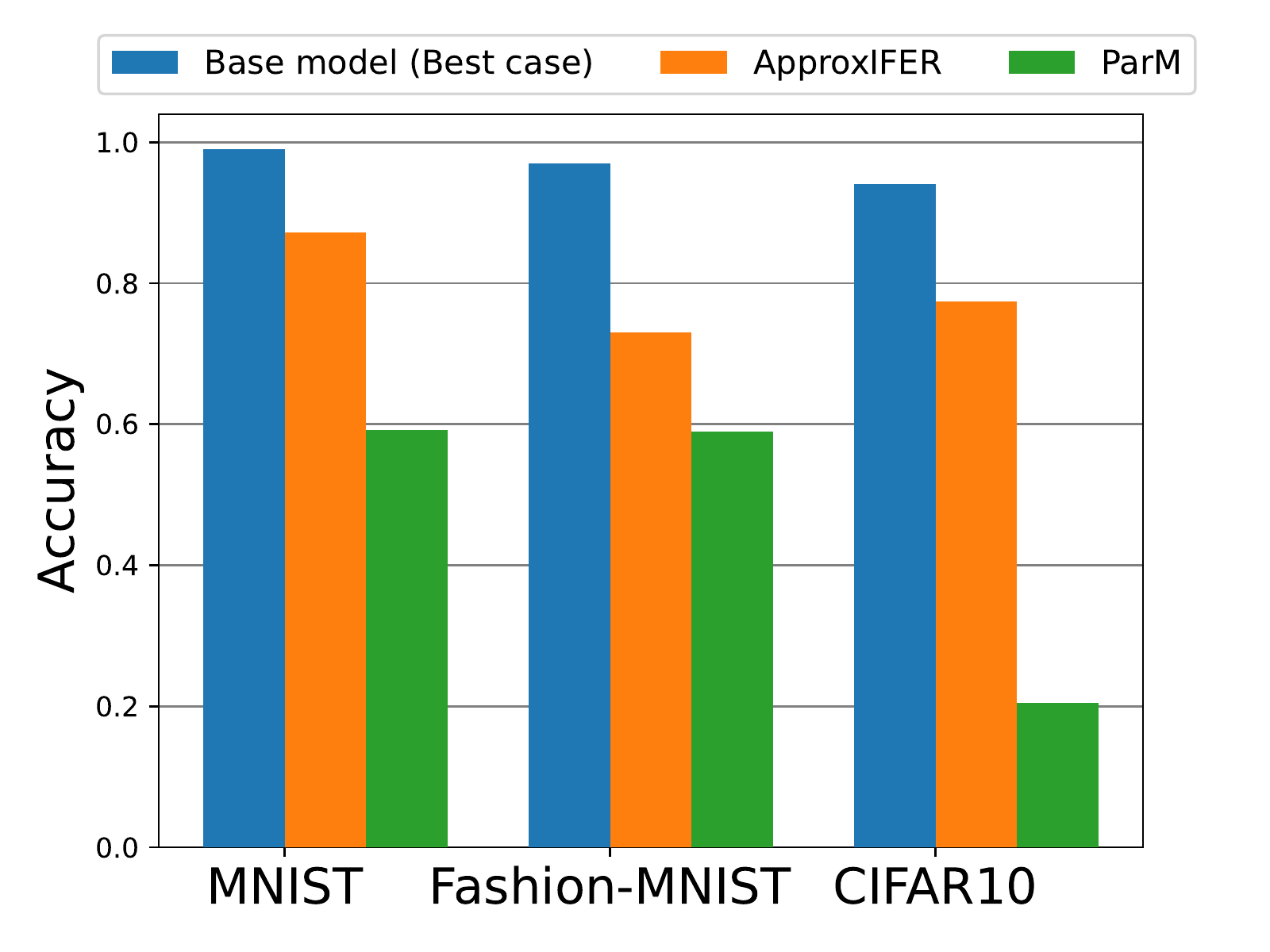}\par\caption{\small Comparison of the accuracy of \namespace with the base model test accuracy for ResNet-$18$ and ParM for  for $K=10$, $S=1$ and $E=0$.  }
\vspace{-2mm}
\label{fig:ResNet}
\end{figure}

These parity models, however, depend on the model hosted by the cloud and are suitable for tolerating one straggler and handling a small number of queries (typically less than $4$). Moreover, they require retraining whenever they are used with a new cloud model. In this work, we take a different approach leveraging approximate coded computing techniques \cite{jahani2020berrut} to design scalable, straggler-resilient and Byzantine-robust prediction serving systems. Our approach relies on rational interpolation techniques \cite{berrut1988rational} to estimate the predictions of the slow and erroneous workers from the predictions of the other workers. Our contributions in this work are summarized as follows. 
\begin{enumerate}

    \item We propose \name, a model-agnostic inference framework that leverages approximate computing techniques. In \name, all workers deploy instances of the model hosted by the cloud and no additional models are required as shown in Fig. \ref{fig:ApproxIferExample}. Furthermore, the encoding and the decoding procedures of \namespace do not depend on the model depolyed by the cloud. This enables \namespace to be easily applied to any neural network architecture.
    \item \namespace is also robust to erroneous workers that return incorrect predictions either unintentionally or adversarially. To do so, we have proposed an algebraic interpolation algorithm 
    to decode the desired predictions from the erroneous coded predictions. \\ \namespace requires a significantly smaller number of workers than the conventional replication method. More specifically, to tolerate $E$ Byzantine adversarial workers, \namespace requires only $2K+2E$ workers whereas the replication-based schemes require $(2E+1)K$ workers. Moreover, \namespace can be set to tolerate any number of stragglers $S$ and errors $E$ efficiently while scaling well with the number of queries $K$, whereas the prior works focused on the case where $S=1, E=0$ and $K=2,3,4$.
    
    \item We run extensive experiments on MNIST, Fashion-MNIST,  and CIFAR-$10$ datasets on VGG, ResNet, DenseNet, and GoogLeNet architectures which show that \namespace improves the prediction accuracy by up to  $58\%$ compared to the prior approaches for large $K$. The results of one of our experiments on ResNet are shown in Fig.~\ref{fig:ResNet}, but we later report extensive experiments on those datasets and architectures showing a consistent significant accuracy improvement over the prior works. 
\end{enumerate}

\textbf{Organization.} The rest of this paper is organized as follows. We describe the problem setting in Section \ref{sec:setting}. Then, we describe \namespace in Section \ref{sec:proposed}. In Section \ref{sec:experiments}, we present our extensive experimental results. In Section \ref{sec:relatedwork}, we discuss the closely-related works. Finally, we discuss some concluding remarks and the future research directions in Section \ref{sec:conclusions}.

\section{Problem Setting}
\label{sec:setting}

\begin{figure}[htb!]
    \centering
    \includegraphics[scale=0.55]{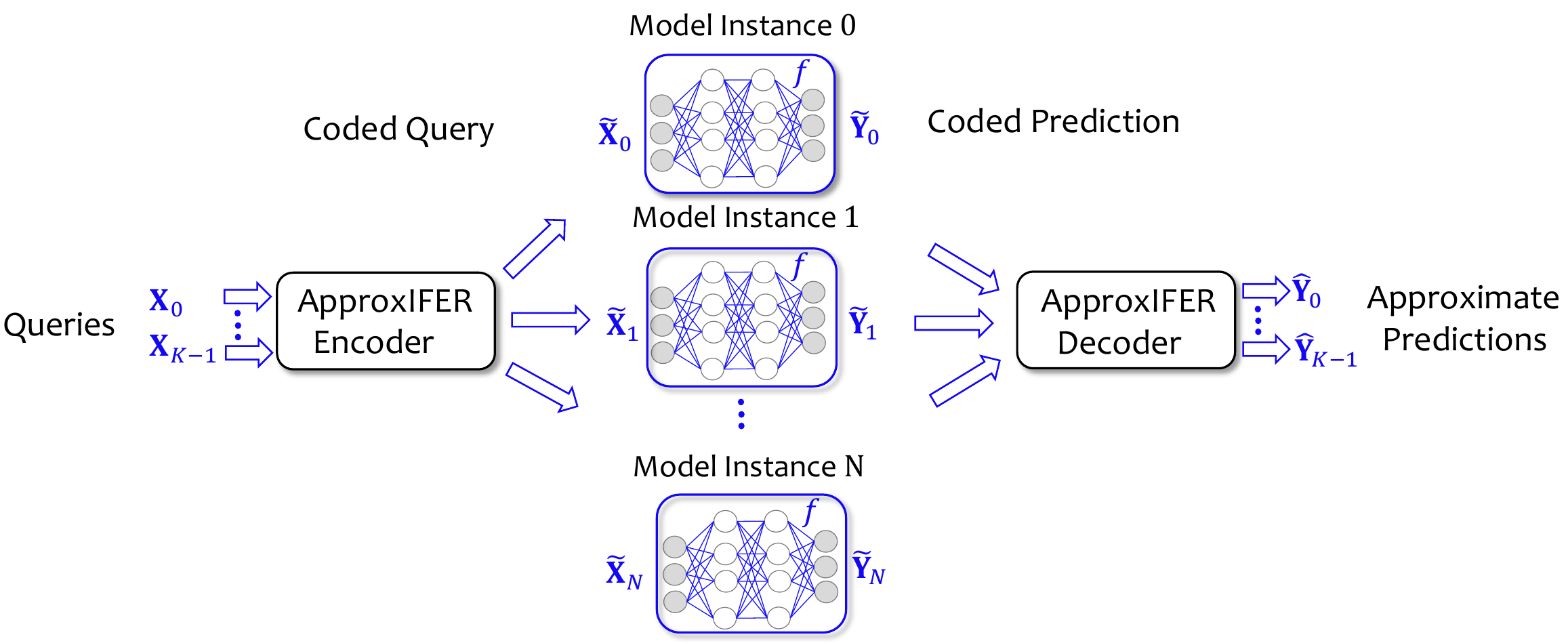}
    \caption{\footnotesize An illustration of the architecture of \name. In \name, all workers have the model deployed by the system $f$, no parity models are required and only an encoder and a decoder are added on top of the conventional replication-based prediction serving systems. The $K$ input queries $\mathbf X_0, \cdots, \mathbf X_{K-1}$ are first encoded. The predictions are then performed on the coded queries. Finally, the approximate predictions $\hat{\mathbf Y}_0, \cdots, \hat{\mathbf Y}_{K-1}$ are recovered from the fastest workers. }
    \label{fig:OurWork}
\end{figure}

\textbf{System Architecture.} We consider a prediction serving system with $N+1$ workers. The prediction serving system is hosting a machine learning model denoted by $f$. We refer to this model as the hosted or the deployed model. Unlike ParM \cite{kosaian2019parity}, all workers have the same model $f$ in our work as shown in Fig. \ref{fig:OurWork}. 

The input queries are grouped such that each group has $K$ queries. We denote the set of $K$ queries in a group by $\mathbf X_0, \mathbf X_1, \cdots, \mathbf X_{K-1}$.

\textbf{Goal.} The goal is to compute the predictions $\mathbf Y_0=f(\mathbf X_0), \mathbf Y_1=f(\mathbf X_1), \cdots, \mathbf Y_{K-1}=f(\mathbf X_{K-1})$ while being resilient to any $S$ stragglers and robust to any $E$ Byzantine workers. 
       
\section{\namespace Algorithm}
\label{sec:proposed}

In this section, we present our proposed protocol based on leveraging approximate coded computing. The encoder and the decoder of \namespace are based on rational functions and rational interpolation. Most coding-theoretic approaches for designing straggler-resilient and Byzantine-robust distributed systems rely on polynomial encoders and polynomial interpolation for decoding. Polynomial interpolation, however, is known to be unstable \cite{berrut2014recent}. On the other hand, rational interpolation is known to be extremely stable and can lead to faster convergence compared to polynomial interpolation \cite{berrut1988rational}. This motivated a recent work to leverage rational functions rather than polynomials to design straggler-resilient distributed training algorithms \cite{jahani2020berrut}. \namespace also leverages rational functions and rational interpolation. We provide a very brief background about rational functions next. 

\textbf{Rational Interpolation Background.} Consider a function $f$, $n+1$ distinct points $a \leq x_0< x_1, \cdots <x_n \leq b$ and the corresponding evaluations of $f$ at these points denoted by $f_0, f_1, \cdots, f_{n-1}, f_n$. Berrut's rational interpolant of $f$ is then defined as follows \cite{berrut1988rational}:
\begin{align}
\label{rational_interpolation}
    r(x) \deff \sum\limits_{i=0}^{n} f_i \ell_i(x),
\end{align}
where $\ell_i(x)$, for $i \in [n]$, where $[n]\,\deff\,\{0,1,2,\dots,n\}$, are the basis functions defined as follows 
\begin{align}
    \ell_i(x) \deff \frac{(-1)^i}{(x-x_i)}\big / \sum\limits_{i=0}^{n} \frac{(-1)^i}{(x-x_i)},
\end{align}
for $i \in [n]$. 
Berrut's rational interpolant has several useful properties as it has no pole on the real line \cite{berrut1988rational} and it is  extremely well-conditioned \cite{bos2011lebesgue,bos2013bounding}. It also converges with rate $O(h)$, where $h \deff \max\limits_{0 \leq i \leq n-1} x_{i+1}-x_{i}$ \cite{floater2007barycentric}.  

\textbf{Rational Interpolation with Erroneous Evaluations.} We now provide our proposed error-locator algorithm for rational interpolation in the presence of Byzantine errors. All the details on how this method works along with the theoretical guarantee and proofs are moved to the Appendix\,\ref{Theory} due to space limitations. Let $\Aavail$ denote the set of indices corresponding to $N-S+1$ available evaluations of $r(x)$ over $x_i$'s, for some $S \geq 1$ that denote the number of stragglers in the context of our system model. Let $\Aadv$, with $|\Aadv|\leq E$, denote the set of indices corresponding to erroneous evaluations. For $i \in \Aavail$, let also $y_i$ denote the available and possibly erroneous evaluation of $r(x)$ at $x_i$. Then we have $y_i = r(x_i)$, for at least $N-S-E+1$ indices $i \in \Aavail$. 
The proposed  algorithm is mainly inspired by the well-known Berlekamp–Welch (BW) decoding algorithm for Reed-Solomon codes in coding theory \cite{blahut2008algebraic}. We tailor the BW algorithm to get a practical algorithm for rational functions that overcomes the numerical issues arising from inevitable round-off errors in the implementation. This algorithm is provided below.   

\begin{algorithm}[H]
 \textbf{Input:}  $x_i$'s, $y_i$'s for $i  \in \Aavail$, $E$ and $K$.\\
\textbf{Output:} Error locations.
  
 \textbf{Step 1}:  Find   polynomials $
P(x) \deff \sum\limits_{i=0}^{K+E-1} P_ix^i,
$
$
Q(x) \deff  \sum\limits_{i=0}^{K+E-1} Q_ix^i 
$ by solving the following system of linear equations:
 $$
 P(x_i)=y_iQ(x_i), \quad \quad \forall i \in \Aavail.
 $$
 \textbf{Step 2}: Set  $a_i=Q(x_i)$, $\forall i \in \Aavail$. 
 
 \textbf{Step 3}: Sort $a_i$'s with respect to their absolute values, i.e., $|a_{i_1}| \leq |a_{i_2}| \leq \cdots |a_{i_{N-S+1}}|$.
 
 \textbf{Return}: $i_1, \cdots, i_{E}.$
\caption{Error-locator algorithm.}
\label{Error_locator}
\end{algorithm}

Note that the equations in Step\,1 of Algorithm\,\ref{Error_locator}  form a homogeneous system of linear equations with $2(K+E)$ unknown variables where  the number of equations is $N-S+1$.  In order to guarantee the existence of a non-trivial solution, we must have
\begin{equation}\label{decoding_condition}
    N\geq 2K+2E+S-1.
\end{equation}
This guarantees the existence of a solution to $P(x)$ and $Q(x)$. 

Next, the encoding and decoding algorithms of \name\ are discussed in detail.
 
 \textbf{\namespace Encoding}. The $K$ input queries $\mathbf X_j$, for $j \in [K-1]$, are first encoded into $N+1$ coded queries, denoted by $\tilde{\mathbf X}_i$, for $i \in [N]$, each given to a worker. As mentioned earlier, the aim is to provide resilience against any $S$ straggler workers and robustness against any $E$ Byzantine adversarial workers. When $E=0$, we assume that $N=K+S-1$ which corresponds to an overhead of $\frac{K+S}{K}$. Otherwise, $N=2(K+E)+S-1$ which corresponds to an overhead of $\frac{2(K+E)+S}{K}$. In general, the overhead is defined as the number of workers divided by the number of queries.

To encode the queries, we leverage Berrut's rational interpolant discussed as follows. First, a rational function $u$ is computed in such a way that it passes through the queries. More specifically,
\begin{align}
u(z)= \sum\limits_{j \in [K-1]} \mathbf X_j \ell_j(z),
\end{align}
 where $\ell_j(x)$, for $j \in [K-1]$, are the basis functions defined as follows
\begin{align}
    \ell_j(z)=\frac{(-1)^j}{(z-\alpha_j)} \big/ \sum\limits_{j \in [K-1]}\frac{(-1)^j}{(z-\alpha_j)},
\end{align}
and $\alpha_j$ is selected as a Chebyshev point of the first kind as 
\begin{align}
\alpha_j=\cos  \frac{(2j+1) \pi}{2K}
\end{align}
for all $j \in [K-1].$
The queries are then encoded using this rational function as follows
\begin{align}
    \tilde{\mathbf X}_i \deff u(\beta_i),
\end{align}
where $\beta_i$ is selected as a Chebyshev point of the second kind as follows 
\begin{align}\label{betas}
\beta_i=\cos  \frac{i\pi}{N},
\end{align}
for $i \in [N]$. The $i$-th worker is then required to compute the prediction on the coded query $\tilde{\mathbf X}_i$. That is, the $i$-th worker computes 
\begin{align}
\tilde{\mathbf Y}_i \deff f(\tilde{\mathbf X}_i)=f(u(\beta_i)),
\end{align}
where $i \in [N]$.\\

\textbf{\namespace Decoding.} When $E=0$, the decoder waits for the results of the fastest $K$ workers before decoding. Otherwise, in the presence of Byzantine workers, i.e., $E>0$, the decoder waits for the results of the fastest $2(K+E)$ workers. After receiving the sufficient number of coded predictions, the decoding approach proceeds with the following two steps. 
\begin{enumerate}
    \item \textbf{Locating Adversarial Workers.} In presence of Byzantine workers that return erroneous predictions aiming at altering the inference results or even unintentionally, we utilize Algorithm\,\ref{Error_locator_Approxifer} provided below to locate them. The predictions corresponding to these workers can be then excluded in the decoding step. Algorithm\,\ref{Error_locator_Approxifer} runs our proposed error-locator algorithm for rational interpolation in presence of errors provided in  Algorithm\,\ref{Error_locator}  several times, each time associated to one of the soft labels in the predictions on the coded queries, i.e., $f(\tilde{\mathbf X}_i)$'s. At the end, we decide the error locations based on a majority vote  on all estimates of the error locations. In Algorithm\,\ref{Error_locator_Approxifer}, $f_j(\tilde{\mathbf X}_i)$ denotes the $j$'th coordinate of $f(\tilde{\mathbf X}_i)$ which is the soft label corresponding to class $j$ in the prediction on the coded query  $f(\tilde{\mathbf X}_i)$. Also, $C$ denotes the total number of classes which is equal to the size of $f(\tilde{\mathbf X}_i)$'s.  

\begin{algorithm}[H]
 \textbf{Input:}  $f(\tilde{\mathbf X}_i)$'s for $i  \in \Aavail$, $\beta_i$'s as specified in  \eqref{betas}, $K$, C and 
 $E$.\\
\textbf{Output:} The set of indices $\Aadv$ corresponding to malicious workers.
 \\
 \\
 Set $\mathbf{I}=[\mathbf{0}]_{C \times E}.$
 
 \textbf{For} $j=1, \cdots, C$\\
 \vspace{-2mm}
 \quad \  \textbf{Step\,1:}
 Set
$
\label{P}
P(x) \deff \sum\limits_{j=0}^{K+E-1} P_jx^j,
$
and
$
Q(x) \deff \sum\limits_{j=1}^{K+E-1} Q_jx^j +1.
$

\vspace{2mm}
\quad \ \ \textbf{Step\,2:} Solve the system of linear equations provided by   
 $$
 P(\beta_i)=f_j(\tilde{\mathbf X}_i)Q(\beta_i), \quad \quad \forall i \in \Aavail.
 $$
 \qquad \qquad \ \   to find the coefficients  $P_j$'s and $Q_j$'s. 
 
 \quad \ \ \textbf{Step\,3}: Set  $a_i=Q(\beta_i)$, $\forall i \in \Aavail$. 
 
  \quad \ \ \textbf{Step\,4}: Sort $a_i$'s increasingly with respect to their absolute values, i.e., $|a_{i_1}| \leq \cdots \leq |a_{i_{N-S+1}}|$.

 \quad \ \ \textbf{Step\,5}: 
 Set $\mathbf{I}[j,:]=[i_1, \cdots, i_{E}].$
 
 \textbf{end}
 
 \textbf{Return}: $\Aadv$:  The set of $E$ most-frequent elements of $\mathbf{I}$.
\caption{\namespace error-locator algorithm.}
\label{Error_locator_Approxifer}
\end{algorithm}

\item \textbf{Decoding}. After excluding the erroneous workers, the approximate predictions can be then recovered from the results of the workers who returned correct coded predictions whose indices are denoted by $\mathcal F$. Specifically, a rational function $r$ is first constructed as follows 
\begin{align}
    r(z)=\frac{1}{\sum\limits_{i \in \mathcal F}\frac{(-1)^i}{(z-\beta_i)}}\sum\limits_{i \in \mathcal F} \frac{(-1)^i}{(z-\beta_i)} f(\tilde{\mathbf X}_i),
\end{align}
where $|\mathcal F|=K$ when $E=0$ and $|\mathcal F|=2K+E$ otherwise.

The approximate predictions denoted by $\hat{\mathbf Y}_0, \cdots, \hat{\mathbf Y}_{K-1}$ then are recovered as follows
\begin{align}
    \hat{\mathbf Y}_j=r(\alpha_j),
\end{align}
for all $j \in [K-1]$. 
\end{enumerate}

\section{Experiments}
\label{sec:experiments}
In this section, we present several experiments to show the effectiveness of \name. 

\subsection{ Experiment Setup}\label{subsec:setup}
More specifically, we perform extensive experiments on the following dastsests and architectures. All experiments are run using PyTorch \cite{paszke2019pytorch} using a MacBook pro with 3.5 GHz dual-core Intel core i7 CPU. The code for implementing \namespace is provided as a supplementary material.    

\textbf{Datasets.}  We run experiments on MNIST \cite{lecun1998gradient}, Fashion-MNIST \cite{xiao2017fashion} and  CIFAR-$10$ \cite{krizhevsky2009learning} datasets. 

\textbf{Architectures.} We consider the following network architectures:  VGG-$16$ \cite{simonyan2014very}, DenseNet-$161$ \cite{huang2017densely}, ResNet-$18$, ResNet-$50$, ResNet-$152$ \cite{he2016deep}, and GoogLeNet \cite{szegedy2015going}.

Some of these architectures such as ResNet-$18$ and VGG-$11$ have been considered to evaluate the performance of earlier works for distributed inference tasks. However, the underlying parity model parameters considered in such works are model-specific, i.e., they are required to be trained from the scratch every time one considers a different base model. This imposes a significant burden on the applicability of such approaches in practice due to their computational heavy training requirements, especially for cases where more than one models parities are needed. In comparison, \namespace is agnostic to the underlying model, and its encoder and decoder do not depend on the employed network architecture as well as the scheme overhead. This enables us to extend our experiments to more complex state-of-the-art models such as ResNet-$50$, ResNet-$152$, DenseNet-$161$, and  GoogLeNet. 

\textbf{Baselines.} We compare \namespace with the ParM framework \cite{kosaian2019parity} in case of tolerating stragglers only. 
Since we are not aware of any baseline that can handle Byzantine workers in the literature other than the straightforward replication approach, we compare the performance of \namespace with the replication scheme.Since the accuracy of the replication approach is the same as the replication of the base model, we compare the test accuracy of \namespace with that of the base model. 

\textbf{Encoding and Decoding.} We employ the encoding algorithm introduced in Section\,\ref{sec:proposed}. In the case of stragglers only, the decoding algorithm in Section\,\ref{sec:proposed} is used. Otherwise, when some of the workers are Byzantine and return erroneous results, we first locate such workers by utilizing the error-locator algorithm provided in Algorithm\,\ref{Error_locator_Approxifer}, exclude their predictions, and then apply the decoding algorithm in Section\,\ref{sec:proposed} to the correct returned results. 

\textbf{Performance metric.} We compare the accuracy of predictions in \namespace with the base model accuracy on the test dataset. In the case of stragglers only, we also compare our results with the accuracy of ParM. 

\subsection{Performance evaluation}

\begin{figure*}[htb!]
\begin{multicols}{2}
\centering
  \includegraphics[width=.9\linewidth]{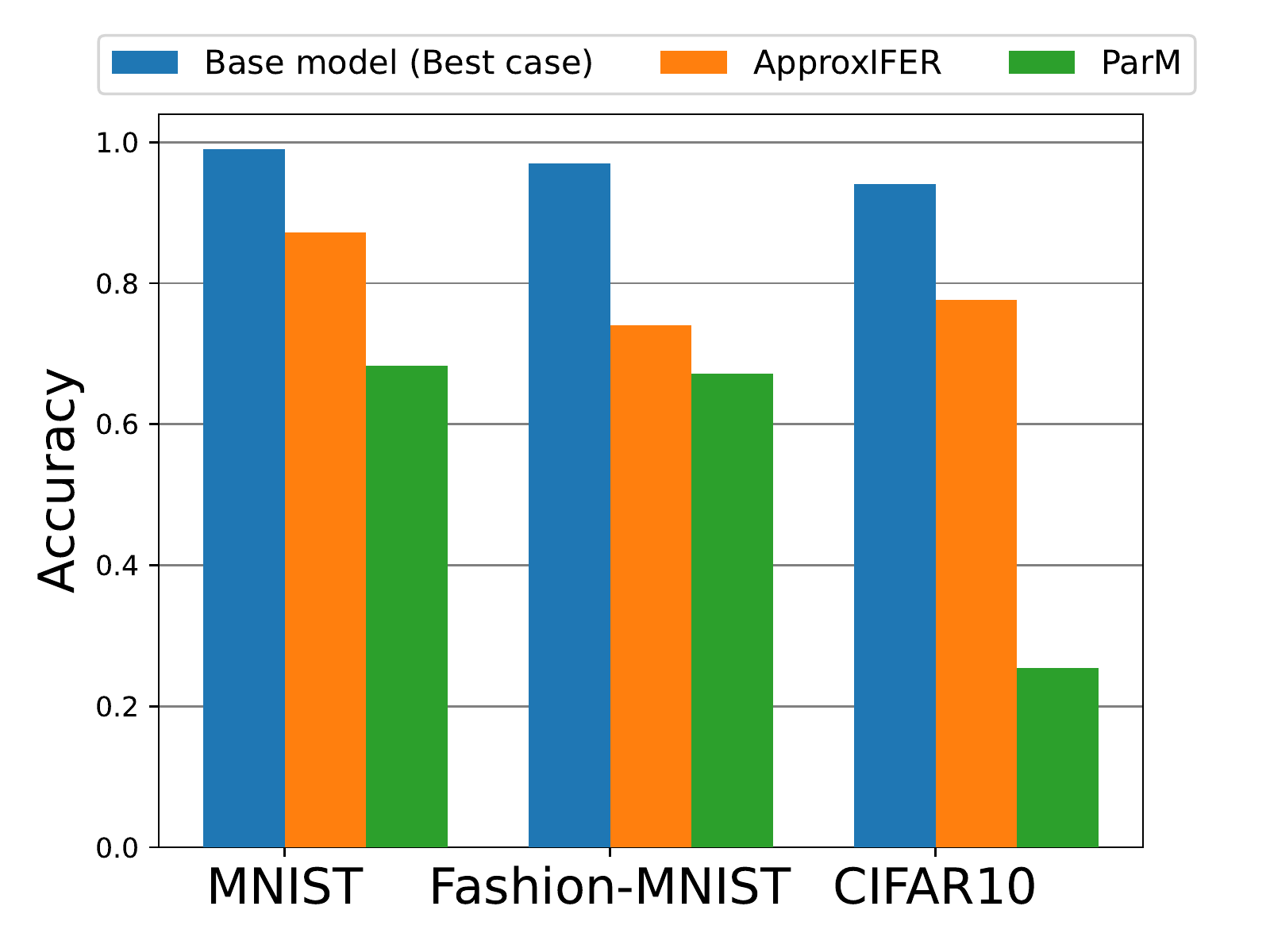}\par\caption{\small Accuracy of \namespace compared with the best case as well as ParM for ResNet-$18$, $K=8$ and $S=1$. }\label{ParM8}
    \includegraphics[width=.9\linewidth]{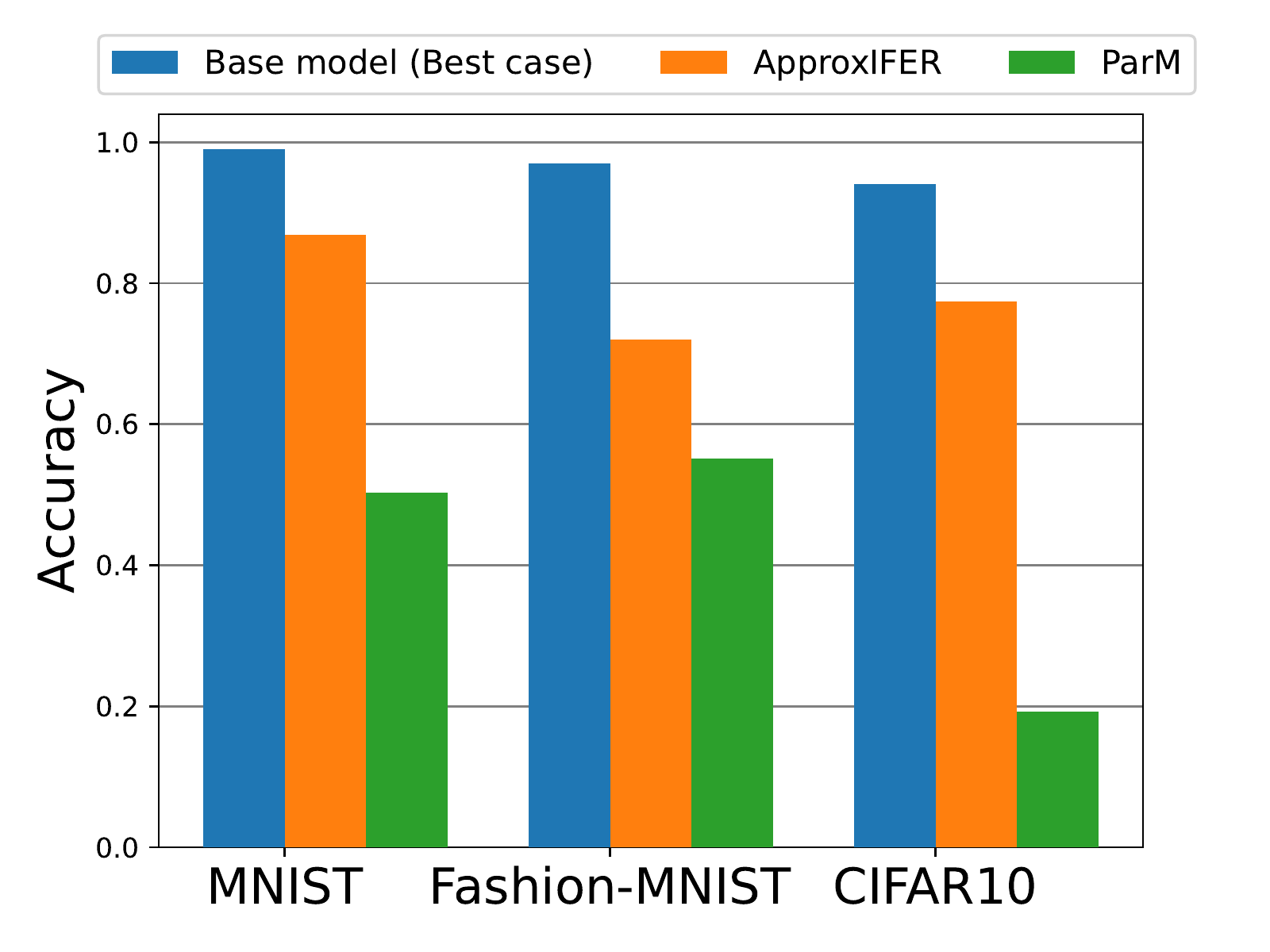}\par\caption{\small Accuracy of \namespace compared with the best case as well as ParM for ResNet-$18$,  $K=12$ and $S=1$.}\label{ParM12}
\end{multicols}
\vspace{-7mm}
\end{figure*}

Our experiments consist of two parts as follows. \\
\textbf{Straggler-Resilience}. In the first part, we consider the case where some of the workers are stragglers and there are no Byzantine workers. We compare our results with  the baseline (ParM) and illustrate that our approach outperforms the baseline results for $K=8,10,12$ and $S=1$. Furthermore, we also illustrate that \namespace can handle multiple stragglers as well by demonstrating its accuracy for $S=2,3$. We then showcase the performance of \namespace over several other more complex architectures for $S=1$ and $K=12$\footnote{The results of ParM are obtained using the codes available at \hyperlink{https://github.com/thesys-lab/parity-models}{https://github.com/thesys-lab/parity-models}.}. To generate the results shown in Figure\,\ref{ParM8} and Figure\,\ref{ParM12} we used pretrained models on CIFAR-$10$ dataset\footnote{The models are available at \hyperlink{https://github.com/huyvnphan/PyTorch_CIFAR10https://github.com/huyvnphan/PyTorch_CIFAR10}{https://github.com/huyvnphan/PyTorch\_ CIFAR10}.}.  

In Figure\,\ref{ParM8} and Figure\,\ref{ParM12}, we compare the performance of \namespace with the base model, i.e., with one worker node and no straggler/Byzantine workers which is also called the best case, as well as with ParM for $K=8$ and $K=12$, respectively, over the test dataset. We considered ResNet-$18$ network architecture and for $K=8$ observed $19\%$, $7\%$ and $51\%$ improvement in the accuracy compared with ParM for the image classification task over MNIST, Fashion-MNIST and CIFAR10 datasets, respectively. For $K=12$, the accuracy improve by $36\%$, $17\%$ and $58\%$, respectively.

\begin{figure*}[htb!]
\begin{multicols}{2}
\centering
  \includegraphics[width=.9\linewidth]{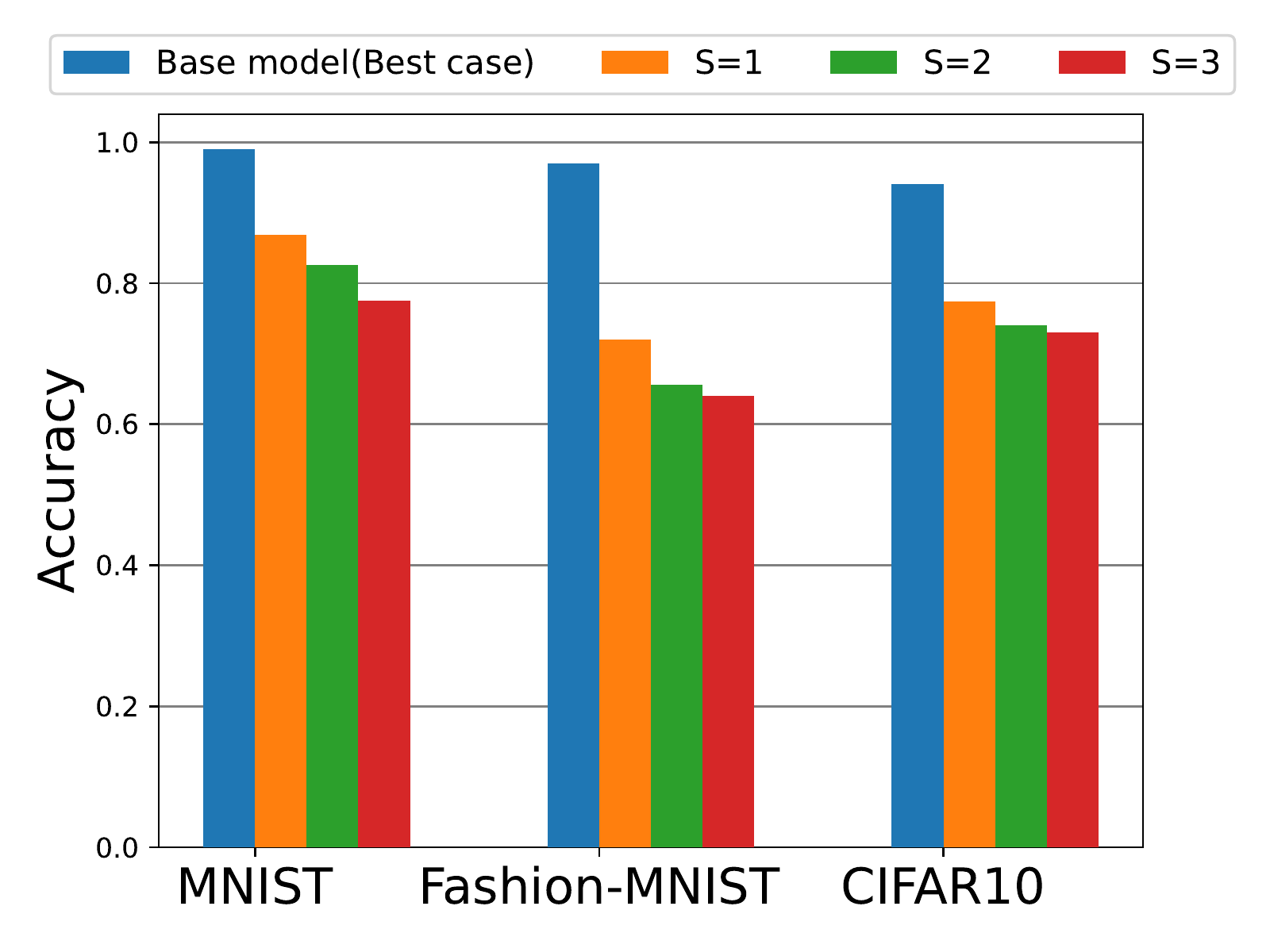}\par\caption{\small Accuracy of \namespace versus the number of stragglers. The network architecture is ResNet-$18$, $K=8$ and $S=1,2,3$.  }\label{Extra_straggler}
    \includegraphics[width=.9\linewidth]{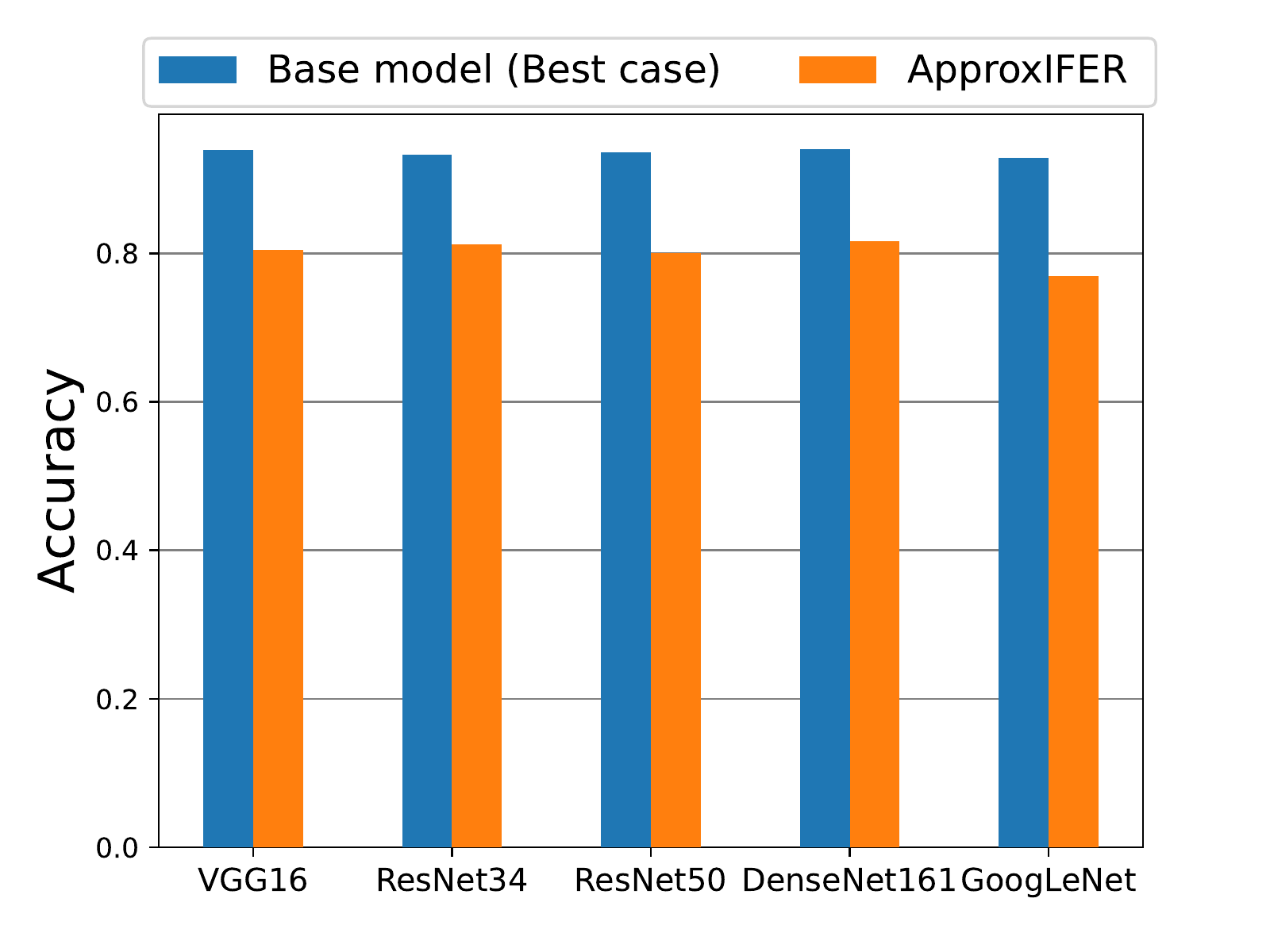}\vspace{0mm}\par\caption{\small Accuracy of \namespace for image classification over CIFAR-$10$ and with various network architectures for $K=8$, $S=1$. }\label{Complex_straggler}
\end{multicols}
\vspace{-7mm}
\end{figure*}

We then extend our experiments by considering more stragglers, e.g., $S=2,3$, and the results are illustrated in Figure\,\ref{Extra_straggler}. The accuracy loss compared with the best case, i.e., no straggler/Byzantine, is not more than $9.4\%$, $8\%$ and $4.4\%$ for MNIST, Fashion-MNIST and CIFAR10 datasets, respectively. Figure\,\ref{Complex_straggler} demonstrates the performance of \namespace for image classification task over CIFAR-$10$ dataset over various state-of-the-art network architectures. The  accuracy loss for $S=1$ compared with the best case is $14\%$, $12\%$, $14\%$, $13\%$ and $16\%$ for   VGG-$16$, ResNet-$34$, ResNet-$50$, DenseNet-$161$ and GoogLeNet, respectively. 

\textbf{Byzantine-Robustness.} In the second part, we provide our experimental results on the performance of \namespace in the presence of Byzantine adversary workers. We include several results for $K=12$ and $E=1,2,3$. In our experiments, the indices of Byzantine workers are determined at random. These workers add a noise that is drawn from a zero-mean normal Gaussian distribution. Lastly, we illustrate that our algorithm performs well for a wide range of standard deviation $\sigma$, namely $\sigma=1,10,100$, thereby demonstrating that our proposed error-locator algorithm performs as promised by the theoretical result regardless of the range of the error values. The results for this experiment are included in Appendix\,\ref{app:sigma_experiments}.


\begin{figure*}[htb!]
\begin{multicols}{2}
\centering
  \includegraphics[width=.9\linewidth]{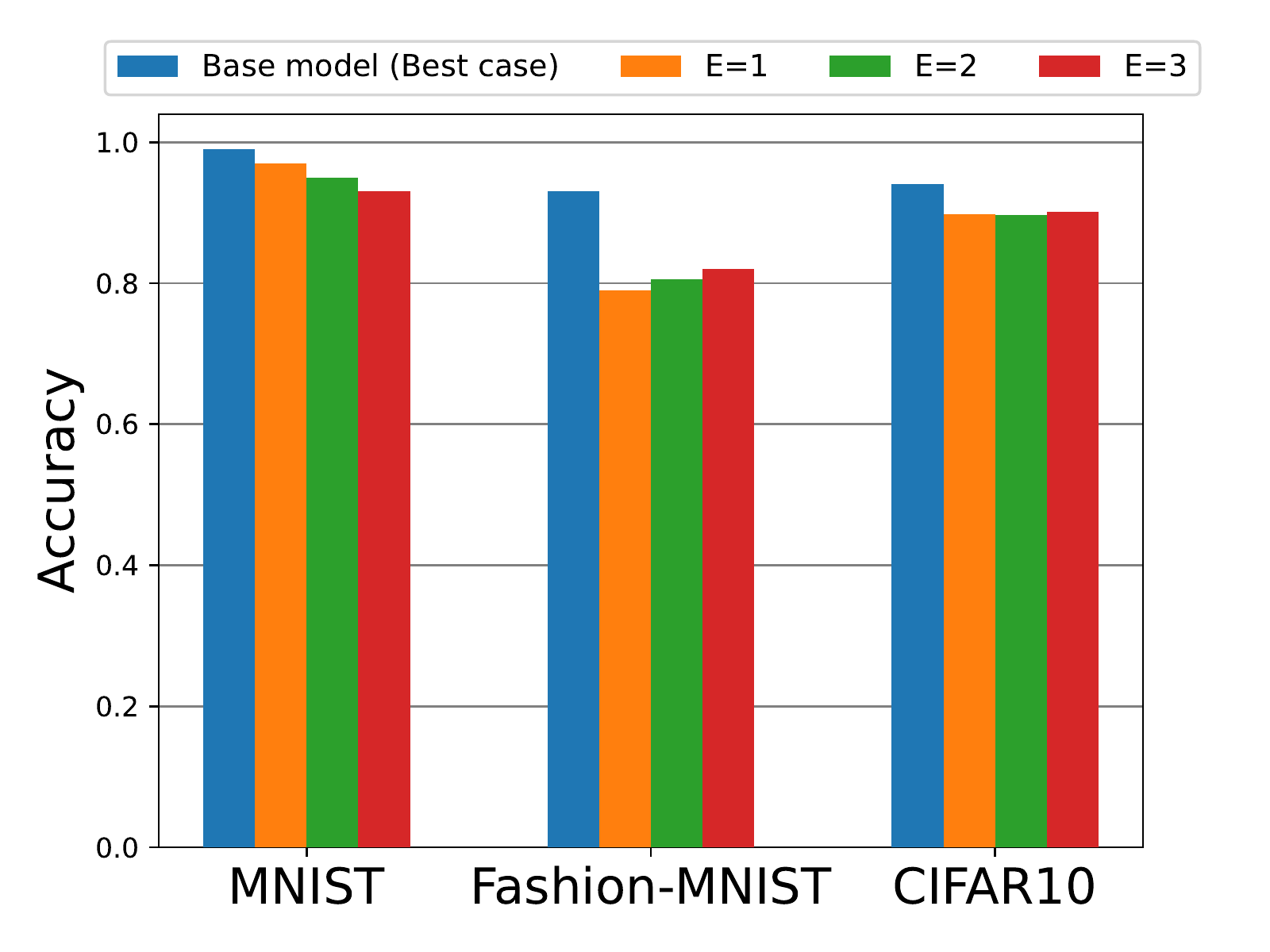}\par\caption{\small Accuracy of \namespace versus the number of errors on ResNet-$18$ for $K=12$, $S=0$, and $E=1,2,3.$   }\label{error_comparison}
    \includegraphics[width=.9\linewidth]{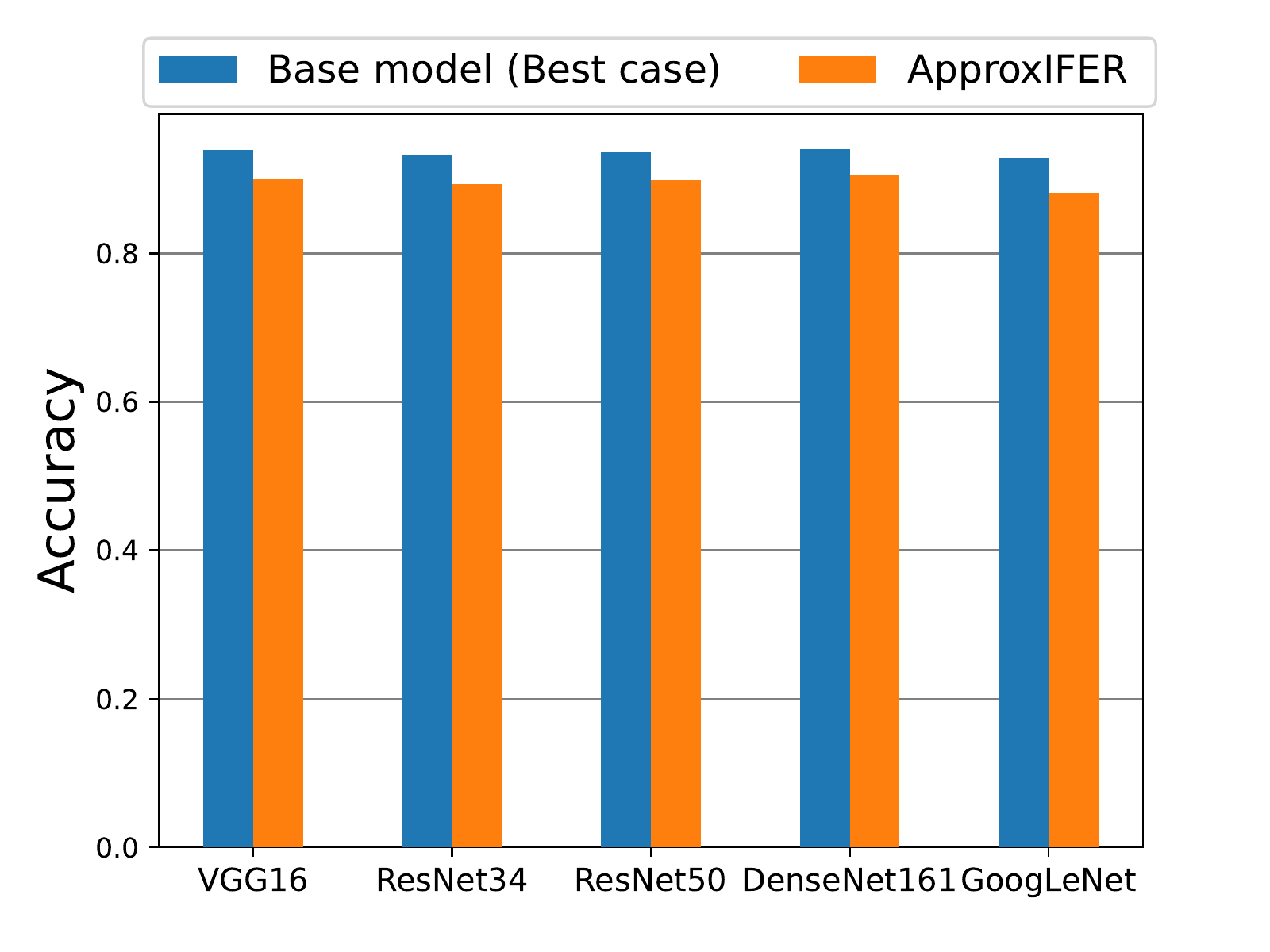}\vspace{0mm}\par\caption{\small Accuracy of \namespace for image classification over CIFAR-$10$ and various network architectures for $K=12$, $S=0$ and $E=2$.}\label{error_complex}
\end{multicols}
\vspace{-7mm}
\end{figure*}

In Figure\,\ref{error_comparison}, we illustrate the accuracy of \namespace with  ResNet-$18$ as the network architecture and for various numbers of Byzantine adversary workers. In this part, we only compare the results with the base model (best case) as there is no other baseline except the straightforward replication scheme. Note that the straightforward replication scheme also attains the best case accuracy, though it requires a significantly higher number of workers, i.e., the number of workers to handle $E$ Byzantine workers in \namespace is $2K+2E$ whereas it is  $(2E+1)K$ in the replication scheme. Our experimental results show that the accuracy loss in \namespace compared with the best case is not more than $6\%$, $4\%$ and $4.2\%$ for MNIST, Fashion-MNIST and CIFAR-$10$ dataset, respectively, for up to $E=3$ malicious workers. These results indicate the success of our proposed algorithm for locating errors in \name, as provided in Algorithm\,\ref{Error_locator_Approxifer}.  
Figure\,\ref{error_complex} demonstrates the accuracy of \namespace in the presence of $E=2$ Byzantine adversaries to perform distributed inference over several underlying network architectures for the CIFAR-$10$ dataset. We observe that the accuracy loss is not more than $5\%$ for VGG-$16$, ResNet-$34$, ResNet-$50$, DenseNet-$161$ and GoogLeNet network architectures when $E=2$. 

\section{Related Works}
\label{sec:relatedwork}
Replication is the most widely used technique for providing straggler resiliency and Byzantine robustness in distributed systems. In this technique, the same task is assigned to multiple workers either proactively or reactively. In the proactive approaches, to tolerate $S$ stragglers, the same task is assigned to $S+1$ workers before starting the computation. While such approaches reduce the latency significantly, they incur significant overhead.  The reactive approaches \cite{zaharia2008improving,dean2013tail} avoid such overhead by assigning the same task to other workers only after a deadline is passed as in Hadoop MapReduce \cite{Apache-Hadoop}. This approach also incurs a significant latency cost as it has to wait before reassigning the tasks.


Recently, coding-theoretic approaches have shown great success in mitigating stragglers in distributed computing and machine learning \cite{lee2017speeding,tandon2017gradient,yu2017polynomial,ye2018communication,wang2019erasurehead,soto2019dual,narra2020collage,dutta2016short,wang2019fundamental}. Such ideas have also been extended to not only provide straggler resiliency, but also Byzantine robustness and data privacy. Specifically, the coded computing paradigm has recently emerged by adapting erasure coding based ideas to design straggler-resilient, Byzantine-robust and private distributed computing systems often involving polynomial-type computations \cite{yang2017coded,yu2019lagrange,subramaniam2019collaborative,soleymani2021list,tang2021verifiable,so2020scalable,sohn2020election,mallick2019fast}. 
However, many applications  involve non-polynomial computations such as distributed training and inference of neural networks. 

A natural approach to get around the polynomial limitation is to approximate any non-polynomial computations. This idea has been leveraged to train a logistic regression model in \cite{so2020scalable}.
This approximation approach, however, is not suitable for neural networks as the number of workers needed is proportional to the degree of the function being computed and also the number of queries. 
Motivated by mitigating these limitations, a learning-based approach was proposed in \cite{kosaian2019parity} to tackle these challenges in prediction serving systems. This idea provides the same straggler-resilience as that of  the underlying erasure code, and hence decouples the straggler-resilience guarantee from the computation carried out by the system. 
This is achieved by learning a parity model known as ParM that transforms the coded queries to coded predictions. 

As we discussed, the learning-based approaches do not scale well. This motivates us in this work to explore a different approach based on approximate coded computing \cite{jahani2021codedsketch,jahani2020berrut}. Approximate computing was leveraged before in distributed matrix-matrix multiplication in \cite{gupta2018oversketch}. Moreover, an approximate \emph{coded} computing approach was developed in   \cite{jahani2021codedsketch} for distributed matrix-matrix multiplication. More recently, a numerically stable straggler-resilient approximate coded computing approach has been developed in \cite{jahani2020berrut}. In particular, this approach is not restricted to polynomials and can be used to \emph{approximately} compute arbitrary
functions unlike the conventional coded computing techniques. One of the key features of this approach is that it uses rational functions \cite{berrut2004barycentric} rather than polynomials to introduce coded redundancy which are known to be numerically stable. This approach, however, does not provide robustness against Byzantine workers. Finally, this approach has been leveraged in distributed training of LeNet-$5$ using MNIST dataset \cite{lecun1998gradient} and resulted in an accuracy that is comparable to the replication-based strategy. 

\section{Conclusions}
\label{sec:conclusions}
In this work, we have introduced \name, a model-agnostic straggler-resilient, and Byzantine-robust framework for prediction serving systems. The key idea of \namespace is that it encodes the queries carefully such that the desired predictions can be recovered efficiently in the presence of both stragglers and Byzantine workers. Unlike the learning-based approaches, our approach does not require training any parity models, can be set to tolerate any number of stragglers and Byzantine workers efficiently. Our experiments on the MNIST, the Fashion-MNIST and the CIFAR-$10$ datasets on various architectures such as VGG, ResNet, DenseNet, and GoogLeNet show that \namespace improves the prediction accuracy by up to $58\%$ compared to the learning-based approaches. An interesting future direction is to extend \namespace to be preserve the privacy of data.  


\bibliographystyle{unsrt}  
\bibliography{Paper} 

\onecolumn

\newpage
\appendix
\section{Rational Interpolation in the Presence of Errors}\label{Theory}


In this section, we provide an algebraic method to interpolate a rational function using its evaluations where some of them are erroneous. Let $\Aadv$ denote the set of indices corresponding to erroneous evaluations and $|\Aadv|\leq E$. 
The proposed  algorithm is mainly inspired by the well-known Berlekamp–Welch (BW) decoding algorithm for Reed-Solomon codes in coding theory \cite{blahut2008algebraic}. This algorithm enables polynomial interpolation in the presence of erroneous evaluations. Our proposed algorithm extends the BW algorithm to rational functions. Extending the BW algorithm to interpolate rational functions in presence of erroneous evaluations has been studied in the literature, e.g., see \cite{boyer2014numerical,kaltofen2013cauchy, blackburn1997fast}. However, we propose a simple yet powerful algorithm that guarantees successful recovery under a slightly different conditions. It is worth noting that no assumption is made on the distribution of the error in this setup and the algorithm finds the rational function and the error locations successfully as long as the number of errors is less than a certain threshold. 

Let $N$ denote the total number of evaluation points. Let also $S$ and $E$ denote the number of points over which the evaluations of $f$ are unavailable (erased) and erroneous (corrupted), respectively. Consider the following polynomial:
\begin{equation}
    \label{Error_locator_polynomail }
    \Lambda(x)=\prod_{i \in \Aadv} (1-\frac{x}{x_i}).
\end{equation}
 This polynomial is referred to as the \emph{error-locator} polynomial as its roots are the evaluation points corresponding to the erroneous evaluations. Suppose that the following rational function $r(x)$ is given:
 
 \begin{equation}
  \label{r(x)}
   r(x)=\frac{p_0+p_1x+ \cdots+p_{K-1}x^{K-1}}{q_0+q_1x+ \cdots+q_{K-1}x^{K-1}}.
 \end{equation}
 
 
Let $x_0, \cdots, x_{N}$ denote the evaluation points. Let $\Aavail$ denote the set of indices corresponding to $N-S+1$ available evaluations of $r(x)$ over $x_i$'s, for some $S \geq 1$ that denote the number of stragglers in the context of our system model. For $i \in \Aavail$, let also $y_i$ denote the available and possibly erroneous evaluation of $r(x)$ at $x_i$. Then we have $y_i = r(x_i)$, for at least $N-S-E+1$ indices $i \in \Aavail$. Then we have
\begin{equation}
    \label{characteristic_equation}
    r(x_i)\Lambda(x_i)=y_i \Lambda(x_i), \quad \quad \forall i \in \Aavail,
\end{equation}
which obviously holds for any $i$ with $y_i = r(x_i)$. Otherwise, when the evaluation is erroneous, i.e., $i \in \Aadv$, then we have $\Lambda(x_i)=0$ implying that \eqref{characteristic_equation} still holds. Let 
\begin{equation}
\label{PP}
P(x) \deff p(x) \Lambda(x)=\sum\limits_{i=0}^{K+E-1} P_ix^i,
\end{equation}
and
\begin{equation}
\label{QQ}
Q(x) \deff q(x) \Lambda(x)=  \sum\limits_{i=0}^{K+E-1} Q_ix^i.
\end{equation}
Plugging in \eqref{PP} and \eqref{QQ} into \eqref{characteristic_equation} results in 
\begin{equation}
    \label{PQ_equations}
    P(x_i)=y_iQ(x_i), \quad \quad \forall i \in \Aavail.
\end{equation}
The equations in \eqref{PQ_equations} form a homogeneous system of linear equations whose unknown variables are  $P_0, \cdots, P_{K+E-1}, Q_0, \cdots, Q_{K+E-1}$. Then the number of unknown variables is $2(K+E)$ and the number of equations is $N-S+1$.  In order to guarantee finding a solution to the set of equations provided in \eqref{PQ_equations}  the number of variables we must have
\begin{equation}\label{decoding_condition}
    N\geq 2K+2E+S-1.
\end{equation}
This guarantees the existence of a solution to $P(x)$ and $Q(x)$. since the underlying system of linear equations is homogeneous.
After determining the polynomials $P(x)$ and $Q(x)$, defined in \eqref{PP} and \eqref{QQ}, respectively, the rational function $r(x)$ is determined by dividing $P(x)$ by $Q(x)$, i.e., $r(x)=\frac{p(x)}{q(x)}=\frac{P(x)}{Q(x)}$. We summarize the proposed algorithm in Algorithm\,\ref{BW-decoder}.

\begin{algorithm}[H]
 \textbf{Input:}  $x_i$'s, $y_i$'s for $i  \in \Aavail$ and $K$.\\
\textbf{Output:} The rational function $r(x)$.
  
 \textbf{Step 1}:  Find the  polynomials $P(x)$ and $Q(x)$, as described in \eqref{PP} and \eqref{QQ}, respectively, such that 
 $$
 P(x_i)=y_iQ(x_i), \quad \quad \forall i \in \Aavail.
 $$
 
 \textbf{Step 2}: Set $r(x)=\frac{P(x)}{Q(x)}.$ 
 
 \textbf{Return}: $r(x).$
\caption{BW-type rational interpolation in presence of errors.}
\label{BW-decoder}
\end{algorithm}

The condition \eqref{decoding_condition} guarantees that the step 1 of Algorithm\,\ref{BW-decoder}  always finds polynomials $P(x)$ and $Q(x)$ successfully. Next, it is shown that $r(x)=\frac{P(x)}{Q(x)}$ in the following theorem.  

\begin{theorem}
\label{thm:errors}
The rational function  returned by Algorithm\,\ref{BW-decoder}  is equal to $r(x)$,  as long as \eqref{decoding_condition} holds and $Q(x)\neq 0$.
\end{theorem}
\begin{proof}
Recall that $r(x)=\frac{p(x)}{q(x)}$. Let $N(x)\deff P(x)q(x)-Q(x)p(x)$. Then, $\deg(N(x))\leq 2K+E-2$. This implies that $N(s)$  has at most $2K+E-1$ roots.

On the other hand, note that $y_i=r(x_i)$ for at least $N-S-E+1$ points since at most $E$  out of $N-S+1$ available evaluations are erroneous. Then, $P(x_i)=r(x_i)Q(x_i)=\frac{p(x_i)}{q(x_i)}Q(x_i)$ which implies $P(x_i)q(x_i)-Q(x_i)p(x_i)=N(x_i)=0$ in at least $N-S-E+1$ points. Hence, $N(x)=0$ if $2K+E-1\leq N-S-E+1$ which implies $r(x)=\frac{p(x)}{q(x)}=\frac{P(x)}{Q(x)}$.  
\end{proof}

Algorithm\,\ref{BW-decoder} is prone to numerical issues in practice due to round-off errors. For implementation purposes, instead of dividing $P(x)$ by $Q(x)$ in step 2 of Algorithm\,\ref{BW-decoder}, we evaluate $Q(x)$ over $x_i$'s for all $i \in \Aavail$ and declare the indices corresponding to $x_i$'s having the least $E$ absolute values as error locations. The rational function $r(x)$ then can be interpolated by excluding the erroneous evaluations. The error-locator algorithm we use in our implementations is provided in Algorithm\,\ref{Error_locator}.

\section{Further Experiments}\label{app:sigma_experiments}
\begin{figure}[H]
\centering
\includegraphics[height=0.145\paperheight, width=0.45\linewidth]{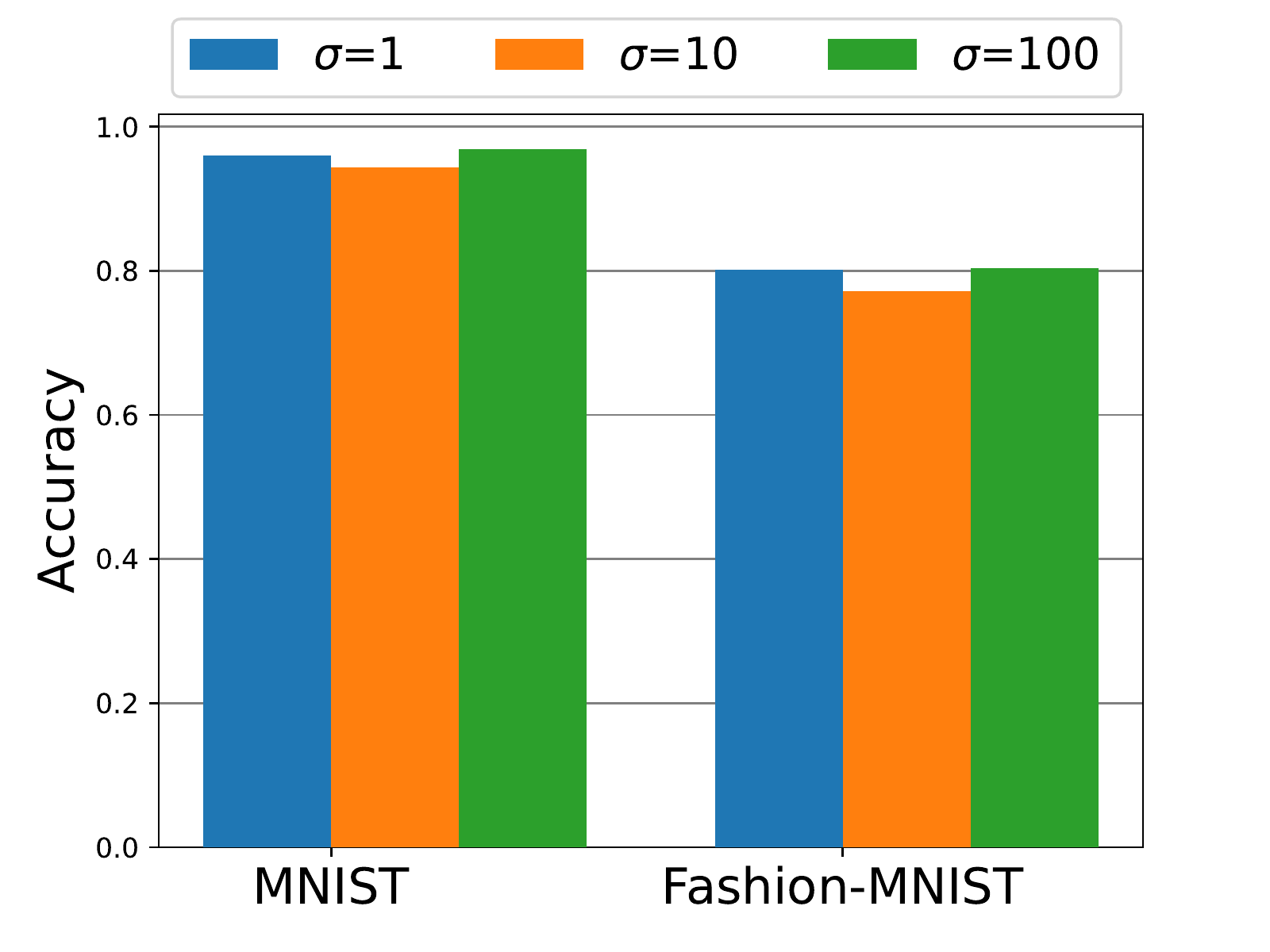}\par\caption{\small Comparison of the accuracy of \namespace for several values of noise standard deviation $\sigma$. Other parameters are $K=8$, $S=0$ and $E=2$. The underlying network architecture is ResNet-$18$ and the experiment is run for MNIST and Fashion-MNIST datasets.  }
\vspace{-2mm}
\label{sigma_error_plot}
\end{figure}

In this section, we illustrate that our algorithm performs well for a wide range of standard deviation $\sigma$.  Our experimental results demonstrate that our proposed error-locator algorithm performs as promised by the theoretical result regardless of the range of the error values. In Figure\,\ref{sigma_error_plot}, we illustrate the accuracy of \namespace with  ResNet-$18$ as the network architecture for the image classification task over MNIST and Fashion-MNIST datasets. The results for $\sigma=1,10,100$ are compared with eachother. It illustrates the proposed error-locator algorithm performs well for a wide range of $\sigma$. 

\section{ Comparison with ParM: Average Case versus Worst Case}

We have compared the accuracy of \namespace with ParM for the worst-case scenario in all reported results in  Section\,\ref{sec:experiments}. For \namespace the worst-case and average-case scenarios are basically the same as all queries are regarded as parity queries. For ParM, the worst-case means that one of  uncoded predictions  is always unavailable. Note that in $\frac{1}{K+1}$ fraction of the times, ParM has access to all uncoded predictions and hence its accuracy is equal to the base model accuracy. In particular, the average-case accuracy of ParM is greater than the worst-case accuracy by at most  $\frac{100}{9}\%\sim11\%$ since $K\geq 8$ in all of our experiments. Hence,  \namespace    still outperforms ParM up to $47\%$ in terms of average-case accuracy of the predictions. 







\end{document}